\documentclass[11pt]{article}
\usepackage[preprint]{acl}
\usepackage{times}
\usepackage{latexsym}
\usepackage[T1]{fontenc}
\usepackage[utf8]{inputenc}
\usepackage{microtype}
\usepackage{amsmath,amssymb}
\usepackage{booktabs}
\usepackage{graphicx}
\usepackage{xcolor}

\newcommand{\new}[1]{#1}
\title{Bridging Teacher--Student Capability Gaps\\with Recurrent Computation}
\author{
  \normalfont
  Zeyi Huang\textsuperscript{1,2} \quad
  Jiuxiang Gu\textsuperscript{1} \quad
  Eli Shechtman\textsuperscript{1} \quad
  Yong Jae Lee\textsuperscript{1,2} \quad
  Zongze Wu\textsuperscript{1} \\[0.5em]
  \textsuperscript{1}Adobe \qquad
  \textsuperscript{2}University of Wisconsin--Madison
}
\begin{document}
\maketitle

\begin{abstract}
Knowledge distillation transfers capabilities from large language models to smaller
students, yet stronger teachers do not always yield better students.
Recent studies report diminishing or even negative returns from teacher
scaling. We hypothesize that these limitations arise in part from
insufficient student capacity to absorb the teacher's knowledge. We therefore
investigate loop and recurrent computation as a way to
expand the student's
effective capacity by modifying its computation graph while keeping the
backbone parameter count fixed.
Under teacher-forced KD during pretraining, weight-shared looping improves performance
but retains teacher-size saturation, whereas cross-token recurrence
avoids the observed reversal with larger teachers.
In mathematical post-training under both teacher-forced KD and on-policy
distillation (OPD), recurrent students scale better with increasing teacher
size than looped and vanilla LLMs do.
Crucially, we show that a pretrained LLM can be
readily fine-tuned into a looped or recurrent architecture jointly
with OPD, without repeating pretraining. Our recurrent Qwen model
improves with larger teachers and consistently outperforms looped and
vanilla LLMs. This
suggests a new post-training paradigm that integrates architectural conversion
into distillation, avoiding a separate conversion stage while substantially
improving student performance and maintaining near-vanilla inference latency.

\end{abstract}

\section{Introduction}
Knowledge distillation (KD) transfers learned capabilities from a teacher
model to a student through teacher supervision
\citep{bucilua2006compression,ba2014deep,hinton2015distilling}.
KD is widely used for model compression, reducing memory
requirements and inference costs for efficient deployment
\citep{romero2015fitnets,mirzadeh2020assistant,gou2021survey,kim2016sequence,sanh2019distilbert,jiao2020tinybert,wang2020minilm}.
Its applications also include transferring capabilities across architectures
\citep{wang2024mamballama}, consolidating ensembles into single models
\citep{hinton2015distilling}, and accelerating diffusion generation through one-
or few-step sampling \citep{salimans2022progressive,song2023consistency}.
In language models, KD provides soft supervision during pretraining
\citep{lu2026strong} and transfers reasoning, coding, and instruction-following
capabilities during post-training \citep{qwen2025report}.
It can also reduce training compute by reusing capabilities
learned by an existing teacher \citep{qwen2025report}.
These applications make KD useful for both efficient deployment
and model development, without requiring a smaller student.

For autoregressive language models, teacher-forced knowledge distillation (TF-KD)
matches student and teacher next-token distributions conditioned on
prefixes from the training data.
In the pretraining formulation studied here, this distillation objective
is combined with cross-entropy on the observed next token,
equivalent to its negative log-likelihood under the student
\citep{lu2026strong,zhang2025capacity,busbridge2025scaling}.
On-policy distillation (OPD) instead provides teacher supervision on
student-generated histories, aligning training with the student's own
generation behavior \citep{agarwal2024opd,gu2023minillm,li2026rethinking}.

A common expectation is that a more capable teacher
will produce a more capable student. However, two recent
studies challenge this expectation in distinct distillation settings.
In teacher-forced pretraining,
\citet{lu2026strong} find that increasing teacher parameters or training tokens can
saturate or reverse distillation gains. In on-policy mathematical post-training,
\citet{li2026rethinking} likewise find that stronger teachers do not guarantee better
students, highlighting thinking-pattern compatibility and genuinely new transferable capabilities.
These observations challenge the assumption that improving the teacher alone
is sufficient to improve knowledge transfer.

We investigate how student
computation affects the ability to benefit from stronger teachers.
We hypothesize that these limitations arise in part from insufficient
student capacity to absorb the teacher's knowledge. Rather than treating
the teacher--student capability gap only as a supervision problem, we
investigate whether changing the student's computation can help it benefit
from stronger teachers. Our central question is: \emph{can recurrent computation
expand the student's effective capacity at fixed backbone parameter count,
and thereby mitigate teacher-size saturation?} Here, effective capacity refers to
the computational flexibility afforded by the student's computation graph, not
an increase in independently parameterized layers. We test this hypothesis
empirically through the student's response to teacher scaling.

We compare weight-shared looping \citep{geiping2025recurrent,zhu2025ouro} and cross-token recurrence
\citep{huang2026lrt} as distinct ways to expand computation at approximately
fixed backbone parameter count.
Looping repeatedly applies the backbone to increase effective depth,
whereas cross-token recurrence carries the previous token's high-level representation
into subsequent computation while retaining a single backbone forward
pass per generated token without additional thinking tokens.

For pretraining, we follow the distillation framework of
\citet{lu2026strong} and compare vanilla, looped, and recurrent students trained
from scratch with teachers ranging from 0.8B to 3.9B.
Figure~\ref{fig:pretraining} reproduces their qualitative finding for vanilla students:
BPB initially decreases but rises beyond intermediate teacher sizes.
Looping lowers BPB but retains this reversal, whereas recurrent
BPB improves overall without increasing as teachers grow larger.

For post-training, we evaluate both TF-KD and OPD on
GSM8K \citep{cobbe2021gsm8k} and MATH500 \citep{hendrycks2021math,lightman2023verify} from our pretraining checkpoints
(Figure~\ref{fig:posttraining}).
In both settings, vanilla gains little from larger teachers,
while looping improves overall; recurrent accuracy increases overall on
GSM8K and monotonically on MATH500.
Recurrence outperforms Loop~3 across all teacher sizes on GSM8K
and with intermediate and larger teachers on MATH500, while
requiring substantially lower decoding latency and a smaller KV
cache at matched sequence length.
These results suggest that recurrence helps students leverage larger
teachers in the pretraining and post-training settings studied here.

Crucially, our approach applies to pretrained Qwen LLMs
(Figure~\ref{fig:qwen-opd}): straightforward fine-tuning jointly converts the backbone into
looped or recurrent architectures and applies OPD, without repeating
pretraining.
Both conversions outperform vanilla OPD on both benchmarks, with
recurrence best across all teacher sizes; these are higher
final accuracies, not uniformly larger incremental OPD gains.
Where a common workflow strengthens a large teacher through
RL before distilling into an unchanged smaller student, we
propose adapting the student's computation to recurrence during OPD.
This route reuses pretrained weights and offers markedly better
student performance with modest inference overhead: in our measured
from-scratch configuration, recurrence adds about 3\% parameters, preserves vanilla
KV-cache size at matched sequence length, and requires one
backbone pass per token at $1.1\times$ vanilla decoding latency.


\section{Related Work}
\paragraph{Teacher strength is not sufficient.}
\citet{lu2026strong} show that increasing teacher parameters or training tokens
can saturate or reverse gains in teacher-forced LLM pretraining distillation.
In a different setting, mathematical post-training OPD, \citet{li2026rethinking} find
that stronger teachers need not yield better students; success depends
on compatible thinking patterns and genuinely new transferable capabilities. These
studies motivate our investigation of whether changing student computation improves
the ability to use stronger teachers across distinct distillation settings.
Larger, more accurate teachers can yield worse students when capacities
are mismatched \citep{cho2019efficacy}. For language models, \citet{zhang2025capacity} report that
optimal teacher size scales approximately linearly with student size in
their studied settings. Teacher assistants bridge this gap through intermediate
networks \citep{mirzadeh2020assistant}. Rather than selecting smaller teachers or introducing
assistants, we vary student computation at fixed backbone scale.

\paragraph{On-policy distillation.}
GKD supervises student-generated sequences with teacher feedback, mitigating the
training--inference prefix mismatch and allowing alternative distribution discrepancies \citep{agarwal2024opd}.
MiniLLM uses reverse KL and on-policy optimization to avoid overestimating
low-probability regions of the teacher distribution \citep{gu2023minillm}. These methods
change the supervision histories or distribution discrepancy; we study the
complementary axis of student computation under teacher-forced and on-policy supervision.

\paragraph{Recurrent and latent computation.}
\emph{Looped computation} reuses shared transformations within a token's computation,
decoupling effective depth from parameter count \citep{dehghani2019universal,lan2019albert,giannou2023programmable}.
Recent work scales this principle through latent iteration, adaptive depth,
and pretrained-model conversion \citep{geiping2025recurrent,zhu2025ouro,bae2025mor,park2026loopus,bae2025recursive}.
Parameter sharing does not remove the cost of repeated computation;
we detail the compute and cache implications in Section~\ref{sec:method}.

\emph{Cross-token recurrence} instead carries already-computed states into subsequent positions.
Upper-to-lower-layer temporal feedback appears in gated-feedback RNNs and Feedback
Transformers, and motivates lightweight latent feedback in LRT
\citep{chung2015gatedfeedback,fan2020feedback,huang2026lrt,huang2026trading}.
Related segment-memory and feedback-attention approaches reuse temporal context
\citep{dai2019transformerxl,hwang2024fam}, but need not implement the same token-level
high-layer-to-early-layer pathway. Additional continuous thought steps offer a separate
axis of computation \citep{hao2025coconut}. We study looping and cross-token
feedback as interventions on distillation compatibility, rather than claiming new
recurrent architectures.

\section{Preliminaries}
\label{sec:method}
We define the student computation patterns and distillation settings used
to study how recurrent computation affects teacher-supervision benefits.

\subsection{Student Computation Patterns}
\paragraph{Looped students.}
A vanilla student processes each token with one backbone pass.
Looped students instead reuse the same backbone across successive passes,
feeding each pass's output into the next to increase effective
depth without adding backbone parameters \citep{geiping2025recurrent,zhu2025ouro}.
With $K$ loops, both training and decoding require $K$ backbone
applications per token. Although weights are shared, each loop produces
distinct KV states; retaining separate caches therefore increases KV-cache storage
approximately $K$-fold. Sharing KV caches across loops can reduce this
overhead \citep{geiping2025recurrent,zhu2025ouro}, but may reduce performance compared
with separate caches \citep{zhu2025ouro}. Our implementation uses independent
KV caches for each loop, without cross-loop cache sharing.
Specifically, we apply the full 12-layer stack $K$ times with
all weights shared, pass the residual stream directly between loops,
and apply the output head only after the final loop.

\paragraph{Cross-token recurrence.}
The recurrent student feeds the previous token's final-layer state
into the current token's computation \citep{huang2026lrt}.
Unlike looping, it reuses an already-computed state across tokens, preserving
one backbone pass per token and approximately vanilla decoding FLOPs,
apart from small feedback overhead. At matched sequence length and
backbone configuration, KV-cache tensor size is unchanged; a small recurrent
state is stored separately. Parallel initialization-and-refinement training adds computation because
previous-token high-level states are unavailable in an ordinary parallel forward
pass
\citep{huang2026lrt}. Thus, extra training work need not increase decoding
passes; this comparison does not establish equal prefill cost or
wall-clock latency.

The fed-back state enters every layer through two lightweight
paths: it is added to the residual stream with a
learnable scale, and per-layer linear maps project it into attention
key and value space, where learned per-head gates mix it
with the local keys and values. The state is zero
at the first position of each sequence. Training approximates the
sequential recurrence with three parallel teacher-forced passes that share weights:
the first pass receives zero feedback, and each later pass
receives the previous pass's final states shifted right by one
position, with gradients flowing through all passes. Cross-entropy is applied
on every pass and the KD term on the last
pass only. Decoding uses a single stream with one forward
pass per token and an ordinary KV cache.

\paragraph{An additional thinking step.}
As an auxiliary pretraining variant, we insert one non-emitting latent
position after each vocabulary token and predict from its final
state \citep{hao2025coconut}. This uses two backbone passes per output
token and increases KV-cache length and FLOPs through additional positions.

A latent position has no token embedding; its input is
the fed-back final state of its paired vocabulary token, and
the next-token prediction and both losses are read there. Latent
positions do not count toward the training token budget.

\subsection{Distillation Settings}
\paragraph{Teacher-forced supervision.}
For a real-data token $y$ with history $c$, let
$p_S(\cdot\mid c)$ denote the student's next-token distribution. Cross-entropy (CE)
trains the student to predict the observed next token by
minimizing its negative log probability:
\begin{equation*}
\mathcal{L}_{\mathrm{CE}}=
\mathbb{E}_{(c,y)\sim\mathrm{data}}[-\log p_S(y\mid c)].
\end{equation*}
Knowledge distillation (KD) instead supervises the student's full next-token
distribution using a fixed teacher. Let $p_T(\cdot\mid c)$ denote
the teacher's distribution. We use forward KL (teacher to student):
\begin{equation*}
\mathcal{L}_{\mathrm{KD}}=
\mathbb{E}_{c\sim\mathrm{data}}\left[
\sum_{v\in\mathcal{V}}p_T(v\mid c)
\log\frac{p_T(v\mid c)}{p_S(v\mid c)}\right].
\end{equation*}
Here, $\mathcal{V}$ is the full vocabulary, and expectations average
over real-data token positions. Both losses use real-data histories,
and both distributions use temperature 1.
Following the pretraining framework of \citet{lu2026strong}, we combine them
with equal weights:
\begin{equation}
\mathcal{L}_{\mathrm{TF}}=0.5\mathcal{L}_{\mathrm{CE}}+0.5\mathcal{L}_{\mathrm{KD}}.
\label{eq:tf}
\end{equation}
The teacher remains fixed throughout student training.

\paragraph{On-policy supervision.}
For a prompt $u$, the current student samples a continuation
$\hat y$.
The fixed teacher evaluates each student-generated history $c_t=(u,\hat y_{<t})$.
Unlike TF-KD, the on-policy KD term uses reverse KL
(student to teacher) on these histories:
\begin{equation}
\begin{aligned}
\mathcal{L}_{\mathrm{OPD}}^{\mathrm{KD}}
&=\mathbb{E}_{c\sim\mathrm{rollout}}\Biggl[
\sum_{v\in\mathcal{V}}p_S(v\mid c)\\
&\qquad\qquad\times\log\frac{p_S(v\mid c)}{p_T(v\mid c)}\Biggr].
\end{aligned}
\label{eq:opd}
\end{equation}
The expectation averages over supervised token positions in student rollouts.
For post-training on-policy updates, we combine this term with
the real-data CE defined above, using equal weights.
CE supervises reference solutions, not sampled tokens. Prompts are
mathematical questions, and sampled continuations are solutions. Post-training alternates
these on-policy updates with TF-KD updates in equal numbers.

\new{Both distributions use temperature 1.
Rollouts are sampled from the current
student at temperature 1 without top-$k$ truncation, used for a
single update, and discarded, so the policy age is zero.
Sampled histories are treated as constants: only the per-position divergence
is differentiated, with no gradient through sampling.
In post-training, the reverse KL is
averaged over sampled answer tokens up to
and including the first end-of-answer token; rollouts that reach the
512-token limit are kept and supervised on all sampled tokens.}

\section{Experiments}
\begin{figure*}[t]
\centering
\includegraphics[width=\textwidth]{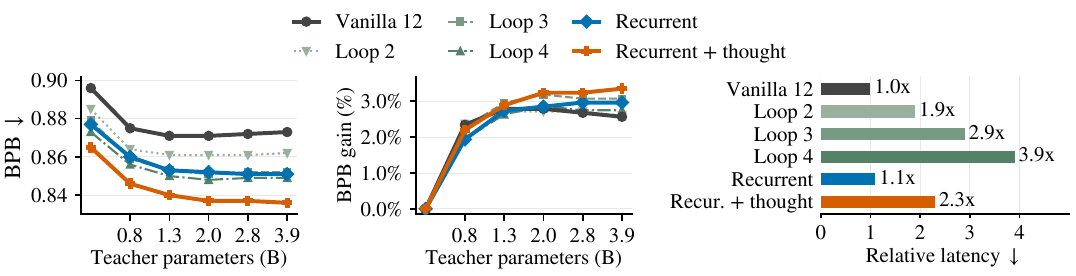}
\caption{TF-KD from scratch on FineWeb-Edu: validation BPB (left)
and relative improvement (center), referenced to each student's
separate CE-only control (unlabeled leftmost point). Right: relative student
latency, normalized to Vanilla~12 (1.0$\times$; lower is better).
Teacher
sizes are categorical. See Appendix Table~\ref{tab:tf} and Equation~\ref{eq:bpb-improvement}.}
\label{fig:pretraining}
\vspace{-10pt}
\end{figure*}

\begin{figure*}[!t]
\centering
\includegraphics[width=\textwidth]{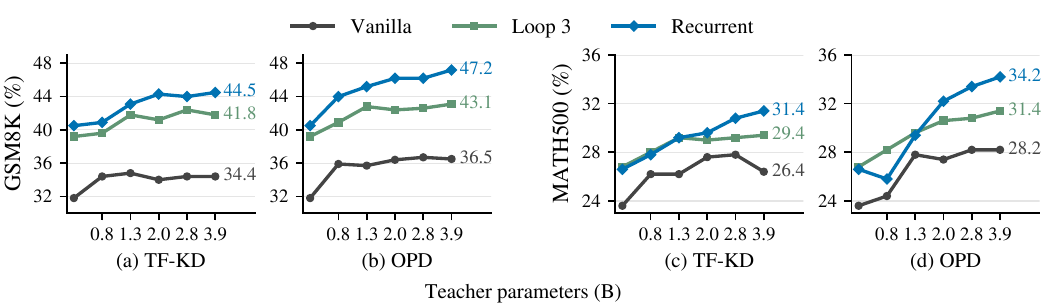}
\caption{Post-training accuracy (\%, $\uparrow$) from our own pretraining weights:
(a) GSM8K TF-KD, (b) GSM8K OPD, (c) MATH500 TF-KD,
and (d) MATH500 OPD. Each unlabeled leftmost point is
the student's no-teacher fine-tuning baseline. Teacher sizes are categorical;
lines connect the baseline and distillation results for comparison.
Full results are in Appendix Table~\ref{tab:post}.}
\label{fig:posttraining}
\vspace{-10pt}
\end{figure*}

We organize the experiments around three questions: teacher-forced pretraining
distillation (Figure~\ref{fig:pretraining}), mathematical reasoning
(Figure~\ref{fig:posttraining}), and adaptation of existing pretrained LLMs
(Figure~\ref{fig:qwen-opd}). Each experiment presents its setting before its results.

\subsection{Teacher-Forced KD from Scratch}
\label{sec:tf-kd-scratch}
\paragraph{Experimental goal.}
We ask whether changing the student's computation graph, while
keeping its backbone parameter count fixed, enables it to
learn more effectively from stronger teachers. We follow the
pretraining distillation framework of \citet{lu2026strong}, \emph{Strong Teacher Not Needed?
On Distillation in LLM Pretraining}. Students are trained from
scratch, not initialized from pretrained checkpoints, using both next-token
cross-entropy (CE) on real text and a knowledge-distillation (KD)
loss against a frozen teacher. We compare vanilla, looped,
and recurrent computation graphs across a teacher sweep, measuring
both absolute held-out BPB and improvement
over each architecture's own CE-only control to separate architectural
gains from teacher-associated benefits.

\paragraph{Setting.}
We first pretrain vanilla autoregressive teachers from scratch on
FineWeb-Edu \citep{penedo2024fineweb} using next-token CE. The teacher sweep
covers approximately 0.8B, 1.3B, 2.0B, 2.8B, and 3.9B parameters.
We then train students from scratch with these teachers
frozen. All students use a 12-layer, approximately 0.4B backbone;
we vary the computation graph across vanilla, looped, and
recurrent architectures. Each student trains for 2,520 steps using
equally weighted CE and forward-KL KD (Equation~\ref{eq:tf}, $\tau=1$), both
on real-text histories. This budget corresponds to 1.32B tokens
and follows scaling-law-guided token budgeting \citep{hoffmann2022training}.
Looped students add no parameters; recurrent
students add 3.2\% feedback-projection overhead, so we fix backbone
parameters rather than total parameters. Appendix~\ref{sec:exp-details} provides full
training and evaluation details and differences from \citet{lu2026strong}.

\paragraph{Controls and evaluation.}
For each architecture, a separate CE-only control matches the
distilled student's initialization seed, data ordering, and training budget.
Comparing these controls reveals performance differences without distillation;
BPB reductions from each architecture's own control measure distillation
gains. Comparing these reductions under the same teacher indicates
whether an architecture gains more from distillation than vanilla.

We evaluate BPB on the first 10.5M tokens of
the held-out FineWeb-Edu shard; lower BPB is better.
Figure~\ref{fig:pretraining}(b) reports the percentage BPB reduction relative to
each architecture's own CE-only baseline, not the vanilla baseline.
We divide each BPB reduction by that architecture's baseline
BPB and express the result as a percentage.

\paragraph{Results: teacher scaling across student computation graphs.}
Figure~\ref{fig:pretraining} shows BPB, improvement over each student's own
CE-only control, and decoding latency; Table~\ref{tab:tf} gives the full
results.
Vanilla reproduces the qualitative finding of \citet{lu2026strong}: a stronger
teacher need not produce a better student. Its CE-only
baseline BPB is 0.896. Although teacher BPB improves from
0.837 to 0.702, vanilla reaches its best distilled BPB
of 0.871 with the 1.3B and 2.0B teachers, then
worsens to 0.873 with the 3.9B teacher. Thus, distillation
remains helpful, but increasing teacher strength eventually reverses gains
relative to smaller teachers.

Looped models \citep{geiping2025recurrent,zhu2025ouro} improve the baseline and distilled BPB,
but at substantial decoding cost. Applying the same Transformer $K$
times per token requires $K$ separate per-loop KV caches
(Section~\ref{sec:method}), increasing cache storage approximately $K$-fold. Measured latency
is $1.9\times$, $2.9\times$, and $3.9\times$ the vanilla latency for
Loop~2, Loop~3, and Loop~4, respectively.
Their CE-only BPB improves to 0.885, 0.879, and 0.873.
With the largest teacher, Loop~3 and Loop~4 also achieve
larger relative distillation improvements than vanilla: 3.07\% and 2.75\%
versus 2.57\%. However, their best distilled BPB occurs with
the 2.0B teacher (0.851 and 0.848), then worsens to
0.852 and 0.849. Extra looping therefore improves student performance
without removing the reversal as teachers become stronger.

Cross-token recurrence \citep{huang2026lrt} provides a different trade-off. Feedback projections
add only 3.2\% parameter overhead, and decoding latency is
$1.1\times$ the vanilla latency, with unchanged KV-cache tensor size
at matched sequence length. Its CE-only BPB is 0.877,
better than
vanilla. With the largest teacher, distillation improves BPB by
2.96\%, also exceeding vanilla's 2.57\%, and reaches 0.851, between
Loop~3's 0.852 and Loop~4's 0.849. Importantly, recurrent BPB decreases
from 0.860 to 0.851 across the teacher sweep, plateauing
for the two largest teachers without the reversal observed
in vanilla and the looped models.

The recurrent model also supports an extra thinking-token mode:
we insert one non-emitting latent position per vocabulary token
\citep{hao2025coconut}. This doubles KV-cache length and requires two backbone
passes per output token, increasing the decoding latency to
$2.3\times$ the vanilla latency. This
variant has the best CE-only baseline BPB (0.865) and,
with the largest teacher, the largest relative distillation improvement
(3.35\%) and lowest final BPB (0.836). Across increasing teacher
sizes, its distilled BPB decreases from 0.846 to 0.836,
with a plateau at 0.837 but no reversal. In
this sweep, recurrence with or without extra thinking exhibits
non-increasing BPB as teachers become stronger; these observed trends
do not establish statistically significant differences between architectures.

\subsection{Distillation on Mathematical Reasoning Tasks}
\paragraph{Setting.}
For Figure~\ref{fig:posttraining}, we initialize post-training distillation from our
own pretraining checkpoints, using the 12-layer, 0.4B student and
16--32-layer, 0.8B--3.9B teacher architectures of Section~\ref{sec:tf-kd-scratch},
pretrained again with twice the token budget to give post-training
a stronger starting point (Appendix~\ref{sec:exp-details}).
This keeps the investigation within one
model family across stages. We compare vanilla, Loop~3, and
recurrent students using final-answer accuracy on GSM8K \citep{cobbe2021gsm8k}
and MATH500 \citep{hendrycks2021math,lightman2023verify}. MATH500 is the 500-problem MATH
test subset introduced by \citet{lightman2023verify}. Figure~\ref{fig:posttraining} shows both
TF-KD and OPD; the main discussion focuses on OPD,
and Appendix Table~\ref{tab:post} provides all numerical results. This sweep
evaluates recurrence without an additional thinking step, which is
evaluated only in pretraining. The reported teacher accuracies come
from the experimental model family, rather than a separate
pretrained-model benchmark table.
\new{\emph{Checkpoints.} Post-training students come from separate pretraining runs
that are longer than those in Figure~\ref{fig:pretraining}.
All three are pretrained without a teacher,
using cross-entropy for 5,040 steps (2.64B tokens). Every post-training run
starts directly from these base checkpoints, with no intervening SFT,
and all teacher conditions of an architecture share the same
student initialization. Teachers are vanilla models pretrained with cross-entropy and
then fine-tuned without a teacher using the same recipe as
the students. We always use the final checkpoint.}

\new{\emph{Training.} All models are fine-tuned for one epoch on the
\texttt{train\_5M} split of OpenMathInstruct-2 \citep{toshniwal2024openmathinstruct2}: 4,746 steps of
524,288 tokens, about 2.49B tokens. There is one joint fine-tuning
run per condition, and GSM8K test and MATH500 are used
only for evaluation. Steps, data, and data order are identical
across student architectures, teacher sizes, TF-KD, OPD, and no-teacher references.
No-teacher controls use the same fine-tuning from the same
student checkpoint, with the teacher terms removed.
TF-KD uses $0.5\,\mathrm{CE}+0.5\,\mathrm{KL}(p_T\|p_S)$ on the reference solutions. OPD alternates
two kinds of steps in equal numbers: a teacher-forced step
with the TF-KD loss above, and an on-policy step with
$0.5\,\mathrm{CE}+0.5\,\mathrm{KL}(p_S\|p_T)$, where the cross-entropy is on the reference solutions
and the reverse KL is computed on 512 solutions
sampled from the
current student.}

\new{\emph{Evaluation.} Teachers and students share one zero-shot protocol with
greedy decoding. GSM8K uses the 1,319 test questions with numeric
exact match. MATH500 uses the 500 problems of the public
\texttt{HuggingFaceH4/MATH-500} test file with Math-Verify equivalence grading. The recurrent
student is decoded recurrently, with one pass per token. Optimization,
rollout, and answer-extraction details are in Appendix~\ref{sec:exp-details}.}

\paragraph{Results: recurrence benefits from larger teachers.}
The results in Figure~\ref{fig:posttraining} show a clearer difference in
teacher scaling. Teacher accuracy increases from 40.1\% to 51.8\%
on GSM8K and from 25.0\% to 34.8\% on MATH500.
On GSM8K, vanilla stays between 35.7\% and 36.7\% for
all five teachers, whereas recurrence rises from 44.0\% to
47.2\%, with a tie at 46.2\% for the 2.0B
and 2.8B teachers. On MATH500, vanilla improves from 24.4\%
to 28.2\% but is unchanged between the 2.8B and
3.9B teachers, while recurrence rises monotonically from 25.8\% to
34.2\%. Loop~3 lies between vanilla and recurrence on GSM8K
at every teacher size and on MATH500 from the
2.0B teacher onward; with the two smallest teachers on
MATH500, it slightly outperforms recurrence. With the 3.9B teacher,
OPD exceeds TF-KD for all three students on both
benchmarks (Appendix Table~\ref{tab:post}).

Crucially, this result concerns the \emph{response to teacher scaling},
not only higher student accuracy. From 2.0B to 3.9B,
vanilla gains 0.1 pp on GSM8K and 0.8 pp
on MATH500, while recurrence gains 1.0 and 2.0 pp.
The recurrent advantage over vanilla rises from 9.8 to
10.7 pp on GSM8K and from 4.8 to 6.0
pp on MATH500 over that range. The GSM8K gains
from 2.0B to 3.9B are about one point, so
MATH500 and the contrast with the vanilla plateau provide
the clearer evidence that cross-token recurrence lets the small
student make better use of larger teachers during mathematical
distillation.

\paragraph{Separating architecture and distillation gains.}
\label{sec:interaction}
In post-training OPD, the teacher-associated advantage generally grows with
teacher size. Without a teacher, recurrence leads vanilla by
8.7 pp on GSM8K and 3.0 pp on MATH500.
With the 3.9B teacher, its leads are 10.7 and
6.0 pp, giving interactions of +2.0 and +3.0 pp.
Equivalently, vanilla improves by 4.7 pp (GSM8K) and 4.6
pp (MATH500) relative to its reference, while recurrence improves
by 6.7 and 7.6 pp. With the 0.8B teacher,
the interactions are -0.6 and -1.6 pp. The MATH500
advantage increases monotonically (1.4, 1.6, 4.8, 5.2, and 6.0
pp), whereas the GSM8K advantage does not (8.1, 9.5,
9.8, 9.5, and 10.7 pp), and most of the
GSM8K lead is already present without a teacher. The
positive interactions at the largest teachers indicate that the
recurrent advantage is not merely a constant baseline offset.
The no-teacher references are matched fine-tuning controls: they start
from the same checkpoint and use the same data, steps,
and data order, with only the teacher terms removed. The
interactions therefore compare runs that differ only in teacher supervision;
they come from single runs and remain descriptive.

\begin{figure*}[!t]
\centering
\includegraphics[width=0.85\textwidth]{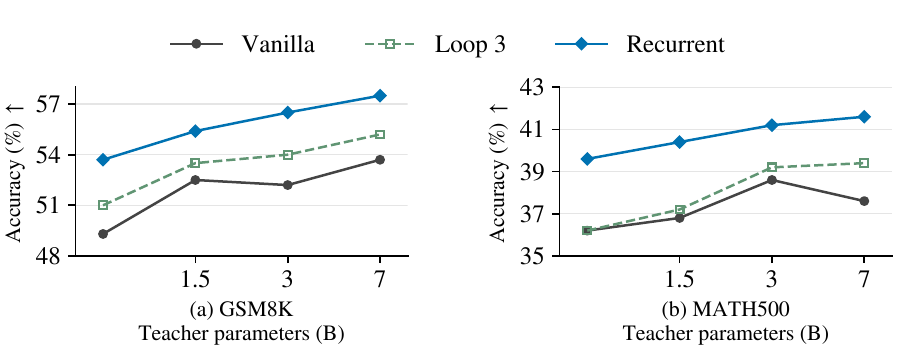}
\caption{Architecture adaptation and OPD from pretrained Qwen2.5-0.5B:
accuracy (\%, $\uparrow$) on (a) GSM8K and (b) MATH500.
Unlabeled leftmost points are no-teacher fine-tuning baselines.
Teacher sizes are categorical. Full results: Appendix Table~\ref{tab:qwen-opd}.}
\label{fig:qwen-opd}
\vspace{-10pt}
\end{figure*}

\subsection{Adapting Pretrained LLMs to Looping and Recurrence}
\paragraph{Setting.}
Rather than training the backbone from scratch, we next
start from pretrained Qwen2.5-0.5B and fine-tune it into Loop~3
or Recurrent architectures (Figure~\ref{fig:qwen-opd}; full results in Appendix
Table~\ref{tab:qwen-opd}). This tests whether recurrent computation can also be
introduced into an existing pretrained LLM.
\new{All runs start from pretrained Qwen2.5-0.5B and fine-tune
the model on OpenMathInstruct-2 \texttt{train\_5M} for one epoch
(4,384 steps, about 2.3B tokens).
The \emph{No teacher} column reports cross-entropy fine-tuning
without a teacher: Vanilla keeps the original architecture, Loop~3 applies
the 24-layer stack three times with shared weights, and
Recurrent adds zero-initialized feedback projections, so that it matches the
pretrained model at initialization, and trains with three parallel passes.
Each OPD run also starts from pretrained Qwen2.5-0.5B and
uses the same data and schedule with a vanilla Qwen2.5
teacher fine-tuned on the same data; the architecture conversion and
distillation therefore happen in the same run. OPD alternates
two kinds of steps in equal numbers: a teacher-forced step
with $0.5\,\mathrm{CE}+0.5\,\mathrm{KL}(p_T\|p_S)$ on reference solutions, and an on-policy step
with $0.5\,\mathrm{CE}+0.5\,\mathrm{KL}(p_S\|p_T)$, where the reverse KL is on 512
sampled student solutions. Unlike the preceding reasoning experiment,
the students are
not pretrained in their target architecture, and
optimization uses AdamW.}

\paragraph{Converting a pretrained model.}
\new{For the recurrent student, the fed-back final state is
normalized to unit RMS before injection, because its scale is
far larger than that of the input embeddings, and
the new feedback parameters use a larger learning rate than
the pretrained weights ($10^{-4}$ versus $2\times10^{-5}$). Loop~3 feeds the output
of the stack back as its input without new
parameters. On 8 H100 GPUs, a training step takes about
1.7$\times$ (Loop~3) and 2.3$\times$ (Recurrent) the time of a
Vanilla step. Details are in Appendix~\ref{sec:exp-details}.}

\paragraph{Fine-tuning instead of repeated pretraining.}
The conversion reuses the pretrained backbone and introduces looping
or recurrence through fine-tuning rather than training a new
LLM from scratch. This avoids the cost of repeating
full pretraining and makes architectural adaptation a practical post-training
route. Avoiding pretraining does not imply negligible fine-tuning cost or
equal inference cost across architectures.

\paragraph{Results: joint architecture adaptation and OPD.}
Without a teacher, Vanilla, Loop~3, and Recurrent score 49.3/36.2,
51.0/36.2, and 53.7/39.6, respectively, on GSM8K/MATH500. The converted
students match or exceed vanilla fine-tuning under the same
budget. Recurrent already
exceeds Vanilla on both benchmarks without
distillation; Loop~3 improves GSM8K and matches Vanilla on MATH500.

Adding OPD to the same fine-tuning run provides further
gains. With the 3B teacher,
the scores reach 52.2/38.6, 54.0/39.2, and 56.5/41.2 for Vanilla,
Loop~3, and Recurrent, respectively. With the 7B teacher, they
reach 53.7/37.6, 55.2/39.4, and 57.5/41.6.
Recurrent achieves the highest student
accuracy on both benchmarks with every teacher. Both Loop~3
and Recurrent outperform Vanilla with the 3B and 7B
teachers, suggesting that the adapted computation graphs
help students better leverage teacher information. This advantage reflects
architectural adaptation and teacher supervision together, rather than OPD
alone.

Increasing teacher size from 1.5B to 3B improves recurrent
accuracy by 1.1 pp on GSM8K and 0.8 pp
on MATH500; Vanilla changes by -0.3 and +1.8 pp,
respectively. From 3B to 7B, Recurrent changes by +1.0
and +0.4 pp, Loop~3 by +1.2 and +0.2 pp,
and Vanilla by +1.5 and -1.0 pp. Relative to
their own no-teacher runs, the 3B
teacher yields gains of 2.9/2.4 pp for Vanilla, 3.0/3.0
pp for Loop~3, and 2.8/1.6 pp for Recurrent.
Thus, the recurrent accuracy advantage also appears with pretrained
Qwen2.5 initialization, but these results do not establish uniformly
larger OPD gains or stronger teacher scaling across both
benchmarks.

\paragraph{Implications for post-training.}
These results suggest a promising new post-training paradigm: convert
a pretrained LLM into a looped or recurrent architecture
and distill it with OPD from a strong teacher
in one fine-tuning run. Architectural
adaptation offers a way to improve the student's use
of teacher knowledge while reusing its pretrained backbone. The
workflow avoids restarting pretraining and achieves better final performance
than vanilla OPD in our reported experiments.

\section{Conclusion}
Teacher quality and student computation should be considered jointly. Looping
and recurrence strengthen small pretrained students, and recurrence with a
thinking step partially mitigates pretraining teacher saturation. When distillation
is applied directly to our pretrained models for mathematical post-training,
recurrent students show non-decreasing accuracy with larger teachers. Vanilla plateaus
on GSM8K across the teacher sweep, while its MATH500 accuracy
improves overall but plateaus between the two largest teachers. With
the 3.9B teacher, recurrence reaches 47.2\% GSM8K and 34.2\% MATH500
accuracy versus 36.5\% and
28.2\% for vanilla. These results motivate
student computation as a practical axis for improving distillation compatibility.
The clearest continued benefits from larger teachers appear with
cross-token recurrence in post-training, suggesting that computational connectivity matters
alongside the amount of computation. Our
experiments do not isolate this interpretation from optimization, calibration, or
training-compute effects.

\section*{Limitations}
Our experiments use one model family, approximately 0.4B-parameter student
backbones, and teachers ranging from 0.8B to 3.9B parameters.
The findings may not generalize to larger students, other architectures,
or other teacher families. Teacher size and pretraining budget vary
jointly, so the sweep does not isolate parameter count alone.
Evaluation covers held-out language modeling and two mathematical benchmarks,
not general downstream capability.
All accuracy differences are reported in percentage points (pp).

\new{All results are single runs, with one seed per configuration
and deterministic greedy evaluation; we report no confidence intervals.}
The reported tables do not quantify training-seed or evaluation uncertainty.
Small BPB differences and larger-teacher accuracy gains should therefore be
interpreted descriptively rather than as established statistically significant effects.
The no-teacher references are matched fine-tuning controls, but with
single runs the teacher-associated gains remain descriptive. Most of the
recurrent advantage on GSM8K already appears
in the reported no-teacher references.

Fixed backbone size does not imply matched training computation.
Looping and recurrent refinement add training work, and recurrence
can introduce lightweight feedback parameters. Our efficiency comparison is
limited to decoding latency; we do not claim matched
training compute. Our decoding
pass counts do not establish equal prefill cost, latency, or
memory use. Finally, we do not directly measure effective capacity
or isolate it from optimization, calibration, and training-compute effects;
the capacity explanation remains a hypothesis, not a demonstrated mechanism.


\section*{Ethical Considerations}
Distillation can transfer undesirable behavior alongside useful capabilities.
Improved language-modeling or mathematical scores do not establish safety,
fairness, or reduced harmful outputs. Recurrent computation is an architectural
intervention, not a safety mechanism. Any release of data, models,
or code should respect the applicable licenses and usage restrictions.
\new{All experiments ran on nodes with 8 H100 GPUs; a
2,520-step student run takes roughly 15--70 minutes depending on architecture
and teacher, and a post-training OPD run takes 3--7 hours.}

\bibliography{references}

@inproceedings{bucilua2006compression,
  title={Model Compression},
  author={Bucilu{\u{a}}, Cristian and Caruana, Rich and Niculescu-Mizil, Alexandru},
  year={2006}, booktitle={Proceedings of the 12th ACM SIGKDD International Conference on Knowledge Discovery and Data Mining},
  doi={10.1145/1150402.1150464}, url={https://doi.org/10.1145/1150402.1150464}
}

@inproceedings{ba2014deep,
  title={Do Deep Nets Really Need to Be Deep?},
  author={Ba, Lei Jimmy and Caruana, Rich},
  year={2014}, booktitle={Advances in Neural Information Processing Systems},
  volume={27}, url={https://arxiv.org/abs/1312.6184}
}

@inproceedings{romero2015fitnets,
  title={{FitNets}: Hints for Thin Deep Nets},
  author={Romero, Adriana and Ballas, Nicolas and Ebrahimi Kahou, Samira and Chassang, Antoine and Gatta, Carlo and Bengio, Yoshua},
  year={2015}, booktitle={International Conference on Learning Representations},
  url={https://arxiv.org/abs/1412.6550}
}

@article{hinton2015distilling,
  title={Distilling the Knowledge in a Neural Network},
  author={Hinton, Geoffrey and Vinyals, Oriol and Dean, Jeff},
  year={2015}, journal={arXiv preprint arXiv:1503.02531},
  doi={10.48550/arXiv.1503.02531}, url={https://arxiv.org/abs/1503.02531}
}

@inproceedings{wang2024mamballama,
  title={The {Mamba} in the {Llama}: Distilling and Accelerating Hybrid Models},
  author={Wang, Junxiong and Paliotta, Daniele and May, Avner and Rush, Alexander M. and Dao, Tri},
  year={2024}, booktitle={Advances in Neural Information Processing Systems},
  url={https://arxiv.org/abs/2408.15237}
}

@inproceedings{salimans2022progressive,
  title={Progressive Distillation for Fast Sampling of Diffusion Models},
  author={Salimans, Tim and Ho, Jonathan},
  year={2022}, booktitle={International Conference on Learning Representations},
  url={https://arxiv.org/abs/2202.00512}
}

@inproceedings{song2023consistency,
  title={Consistency Models},
  author={Song, Yang and Dhariwal, Prafulla and Chen, Mark and Sutskever, Ilya},
  year={2023}, booktitle={Proceedings of the 40th International Conference on Machine Learning},
  url={https://arxiv.org/abs/2303.01469}
}

@article{qwen2025report,
  title={{Qwen3} Technical Report},
  author={{Qwen Team}},
  year={2025}, journal={arXiv preprint arXiv:2505.09388},
  doi={10.48550/arXiv.2505.09388}, url={https://arxiv.org/abs/2505.09388}
}

@article{gou2021survey,
  title={Knowledge Distillation: A Survey},
  author={Gou, Jianping and Yu, Baosheng and Maybank, Stephen John and Tao, Dacheng},
  year={2021}, journal={International Journal of Computer Vision},
  doi={10.1007/s11263-021-01453-z}, url={https://arxiv.org/abs/2006.05525}
}

@inproceedings{kim2016sequence,
  title={Sequence-Level Knowledge Distillation},
  author={Kim, Yoon and Rush, Alexander M.},
  year={2016}, booktitle={Proceedings of the 2016 Conference on Empirical Methods in Natural Language Processing},
  pages={1317--1327}, doi={10.18653/v1/D16-1139},
  url={https://aclanthology.org/D16-1139/}
}

@article{sanh2019distilbert,
  title={{DistilBERT}, a Distilled Version of {BERT}: Smaller, Faster, Cheaper and Lighter},
  author={Sanh, Victor and Debut, Lysandre and Chaumond, Julien and Wolf, Thomas},
  year={2019}, journal={arXiv preprint arXiv:1910.01108},
  doi={10.48550/arXiv.1910.01108}, url={https://arxiv.org/abs/1910.01108}
}

@inproceedings{jiao2020tinybert,
  title={{TinyBERT}: Distilling {BERT} for Natural Language Understanding},
  author={Jiao, Xiaoqi and Yin, Yichun and Shang, Lifeng and Jiang, Xin and Chen, Xiao and Li, Linlin and Wang, Fang and Liu, Qun},
  year={2020}, booktitle={Findings of the Association for Computational Linguistics: EMNLP 2020},
  pages={4163--4174}, doi={10.18653/v1/2020.findings-emnlp.372},
  url={https://aclanthology.org/2020.findings-emnlp.372/}
}

@inproceedings{wang2020minilm,
  title={{MiniLM}: Deep Self-Attention Distillation for Task-Agnostic Compression of Pre-Trained Transformers},
  author={Wang, Wenhui and Wei, Furu and Dong, Li and Bao, Hangbo and Yang, Nan and Zhou, Ming},
  year={2020}, booktitle={Advances in Neural Information Processing Systems},
  volume={33}, pages={5776--5788}, url={https://arxiv.org/abs/2002.10957}
}

@article{busbridge2025scaling,
  title={Distillation Scaling Laws},
  author={Busbridge, Dan and Shidani, Amitis and Weers, Floris and Ramapuram, Jason and Littwin, Etai and Webb, Russ},
  year={2025}, journal={arXiv preprint arXiv:2502.08606},
  doi={10.48550/arXiv.2502.08606}, url={https://arxiv.org/abs/2502.08606}
}

@inproceedings{mirzadeh2020assistant,
  title={Improved Knowledge Distillation via Teacher Assistant},
  author={Mirzadeh, Seyed Iman and Farajtabar, Mehrdad and Li, Ang and Levine, Nir and Matsukawa, Akihiro and Ghasemzadeh, Hassan},
  year={2020}, booktitle={Proceedings of the AAAI Conference on Artificial Intelligence},
  volume={34}, pages={5191--5198},
  doi={10.1609/aaai.v34i04.5963}, url={https://ojs.aaai.org/index.php/AAAI/article/view/5963}
}

@article{lu2026strong,
  title={Strong Teacher Not Needed? On Distillation in {LLM} Pretraining},
  author={Lu, Taiming and Liu, Zhuang}, year={2026},
  journal={arXiv preprint arXiv:2605.23857},
  doi={10.48550/arXiv.2605.23857}, url={https://arxiv.org/abs/2605.23857}
}

@article{penedo2024fineweb,
  title={The {FineWeb} Datasets: Decanting the Web for the Finest Text Data at Scale},
  author={Penedo, Guilherme and Kydl{\'{i}}{\v{c}}ek, Hynek and {Ben Allal}, Loubna and Lozhkov, Anton and Mitchell, Margaret and Raffel, Colin and {von Werra}, Leandro and Wolf, Thomas},
  year={2024}, journal={arXiv preprint arXiv:2406.17557},
  doi={10.48550/arXiv.2406.17557}, url={https://arxiv.org/abs/2406.17557}
}

@article{li2026rethinking,
  title={Rethinking On-Policy Distillation of Large Language Models: Phenomenology, Mechanism, and Recipe},
  author={Li, Yaxuan and Zuo, Yuxin and He, Bingxiang and Zhang, Jinqian and Xiao, Chaojun and Qian, Cheng and Yu, Tianyu and Gao, Huan-ang and Yang, Wenkai and Liu, Zhiyuan and Ding, Ning},
  year={2026}, journal={arXiv preprint arXiv:2604.13016},
  doi={10.48550/arXiv.2604.13016}, url={https://arxiv.org/abs/2604.13016}
}

@inproceedings{agarwal2024opd,
  title={On-Policy Distillation of Language Models: Learning from Self-Generated Mistakes},
  author={Agarwal, Rishabh and Vieillard, Nino and Zhou, Yongchao and Stanczyk, Piotr and Ramos Garea, Sabela and Geist, Matthieu and Bachem, Olivier},
  year={2024}, booktitle={International Conference on Learning Representations},
  url={https://proceedings.iclr.cc/paper_files/paper/2024/hash/5be69a584901a26c521c2b51e40a4c20-Abstract-Conference.html}
}

@article{huang2026lrt,
  title={Latent Recurrent Transformer: Architecture Exploration, Training Strategies, and Scaling Behavior},
  author={Huang, Zeyi and He, Xuehai and Ren, LiLiang and Wang, Yiping and Peng, Baolin and Cheng, Hao and Wang, Shuohang and He, Pengcheng and Gao, Jianfeng and Lee, Yong Jae and Shen, Yelong},
  year={2026}, journal={arXiv preprint arXiv:2605.26797},
  doi={10.48550/arXiv.2605.26797}, url={https://arxiv.org/abs/2605.26797}
}

@inproceedings{geiping2025recurrent,
  title={Scaling up Test-Time Compute with Latent Reasoning: A Recurrent Depth Approach},
  author={Geiping, Jonas and McLeish, Sean and Jain, Neel and Kirchenbauer, John and Singh, Siddharth and Bartoldson, Brian R. and Kailkhura, Bhavya and Bhatele, Abhinav and Goldstein, Tom},
  year={2025}, booktitle={Advances in Neural Information Processing Systems},
  url={https://proceedings.neurips.cc/paper_files/paper/2025/hash/3b01972cf31e6fa0fe29e4b8b5c2a0a1-Abstract-Conference.html}
}

@article{qwen2024report,
  title={{Qwen2.5} Technical Report}, author={{Qwen Team}}, year={2024},
  journal={arXiv preprint arXiv:2412.15115},
  doi={10.48550/arXiv.2412.15115}, url={https://arxiv.org/abs/2412.15115}
}

@article{cobbe2021gsm8k,
  title={Training Verifiers to Solve Math Word Problems},
  author={Cobbe, Karl and Kosaraju, Vineet and Bavarian, Mohammad and Chen, Mark and Jun, Heewoo and Kaiser, Lukasz and Plappert, Matthias and Tworek, Jerry and Hilton, Jacob and Nakano, Reiichiro and Hesse, Christopher and Schulman, John},
  year={2021}, journal={arXiv preprint arXiv:2110.14168},
  doi={10.48550/arXiv.2110.14168}, url={https://arxiv.org/abs/2110.14168}
}

@article{hendrycks2021math,
  title={Measuring Mathematical Problem Solving With the {MATH} Dataset},
  author={Hendrycks, Dan and Burns, Collin and Kadavath, Saurav and Arora, Akul and Basart, Steven and Tang, Eric and Song, Dawn and Steinhardt, Jacob},
  year={2021}, journal={arXiv preprint arXiv:2103.03874},
  doi={10.48550/arXiv.2103.03874}, url={https://arxiv.org/abs/2103.03874}
}

@article{lightman2023verify,
  title={Let's Verify Step by Step},
  author={Lightman, Hunter and Kosaraju, Vineet and Burda, Yura and Edwards, Harri and Baker, Bowen and Lee, Teddy and Leike, Jan and Schulman, John and Sutskever, Ilya and Cobbe, Karl},
  year={2023}, journal={arXiv preprint arXiv:2305.20050},
  doi={10.48550/arXiv.2305.20050}, url={https://arxiv.org/abs/2305.20050}
}

@inproceedings{cho2019efficacy,
  title={On the Efficacy of Knowledge Distillation},
  author={Cho, Jang Hyun and Hariharan, Bharath},
  year={2019}, booktitle={Proceedings of the IEEE/CVF International Conference on Computer Vision},
  pages={4794--4802}, url={https://arxiv.org/abs/1910.01348}
}

@inproceedings{zhang2025capacity,
  title={Towards the Law of Capacity Gap in Distilling Language Models},
  author={Zhang, Chen and Li, Qiuchi and Song, Dawei and Ye, Zheyu and Gao, Yan and Hu, Yan},
  year={2025}, booktitle={Proceedings of the 63rd Annual Meeting of the Association for Computational Linguistics},
  url={https://arxiv.org/abs/2311.07052}
}

@article{gu2023minillm,
  title={{MiniLLM}: On-Policy Distillation of Large Language Models},
  author={Gu, Yuxian and Dong, Li and Wei, Furu and Huang, Minlie},
  year={2023}, journal={arXiv preprint arXiv:2306.08543},
  doi={10.48550/arXiv.2306.08543}, url={https://arxiv.org/abs/2306.08543}
}

@article{zhu2025ouro,
  title={Scaling Latent Reasoning via Looped Language Models},
  author={Zhu, Rui-Jie and Wang, Zixuan and Hua, Kai and Zhang, Tianyu and Li, Ziniu and Que, Haoran and Wei, Boyi and Wen, Zixin and Yin, Fan and others},
  year={2025}, journal={arXiv preprint arXiv:2510.25741},
  doi={10.48550/arXiv.2510.25741}, url={https://arxiv.org/abs/2510.25741}
}

@article{park2026loopus,
  title={{LoopUS}: Recasting Pretrained {LLMs} into Looped Latent Refinement Models},
  author={Park, Taekhyun and Lee, Yongjae and Kim, Dohee and Bae, Hyerim},
  year={2026}, journal={arXiv preprint arXiv:2605.11011},
  doi={10.48550/arXiv.2605.11011}, url={https://arxiv.org/abs/2605.11011}
}

@inproceedings{bae2025recursive,
  title={Relaxed Recursive Transformers: Effective Parameter Sharing with Layer-wise {LoRA}},
  author={Bae, Sangmin and Fisch, Adam and Harutyunyan, Hrayr and Ji, Ziwei and Kim, Seungyeon and Schuster, Tal},
  year={2025}, booktitle={International Conference on Learning Representations},
  url={https://openreview.net/forum?id=WwpYSOkkCt}
}

@inproceedings{hao2025coconut,
  title={Training Large Language Models to Reason in a Continuous Latent Space},
  author={Hao, Shibo and Sukhbaatar, Sainbayar and Su, DiJia and Li, Xian and Hu, Zhiting and Weston, Jason and Tian, Yuandong},
  year={2025}, booktitle={Conference on Language Modeling},
  url={https://arxiv.org/abs/2412.06769}
}

@inproceedings{dehghani2019universal,
  title={Universal Transformers},
  author={Dehghani, Mostafa and Gouws, Stephan and Vinyals, Oriol and Uszkoreit, Jakob and Kaiser, {\L}ukasz},
  year={2019}, booktitle={International Conference on Learning Representations},
  url={https://arxiv.org/abs/1807.03819}
}

@article{lan2019albert,
  title={{ALBERT}: A Lite {BERT} for Self-supervised Learning of Language Representations},
  author={Lan, Zhenzhong and Chen, Mingda and Goodman, Sebastian and Gimpel, Kevin and Sharma, Piyush and Soricut, Radu},
  year={2019}, journal={arXiv preprint arXiv:1909.11942},
  doi={10.48550/arXiv.1909.11942}, url={https://arxiv.org/abs/1909.11942}
}

@article{giannou2023programmable,
  title={Looped Transformers as Programmable Computers},
  author={Giannou, Angeliki and Rajput, Shashank and Sohn, Jy-yong and Lee, Kangwook and Lee, Jason D. and Papailiopoulos, Dimitris},
  year={2023}, journal={arXiv preprint arXiv:2301.13196},
  doi={10.48550/arXiv.2301.13196}, url={https://arxiv.org/abs/2301.13196}
}

@inproceedings{dai2019transformerxl,
  title={{Transformer-XL}: Attentive Language Models Beyond a Fixed-Length Context},
  author={Dai, Zihang and Yang, Zhilin and Yang, Yiming and Carbonell, Jaime and Le, Quoc V. and Salakhutdinov, Ruslan},
  year={2019}, booktitle={Proceedings of the 57th Annual Meeting of the Association for Computational Linguistics},
  url={https://arxiv.org/abs/1901.02860}
}

@article{fan2020feedback,
  title={Addressing Some Limitations of Transformers with Feedback Memory},
  author={Fan, Angela and Lavril, Thibaut and Grave, Edouard and Joulin, Armand and Sukhbaatar, Sainbayar},
  year={2020}, journal={arXiv preprint arXiv:2002.09402},
  doi={10.48550/arXiv.2002.09402}, url={https://arxiv.org/abs/2002.09402}
}

@article{bae2025mor,
  title={Mixture-of-Recursions: Learning Dynamic Recursive Depths for Adaptive Token-Level Computation},
  author={Bae, Sangmin and Kim, Yujin and Bayat, Reza and Kim, Sungnyun and Ha, Jiyoun and Schuster, Tal and Fisch, Adam and Harutyunyan, Hrayr and Ji, Ziwei and Courville, Aaron and Yun, Se-Young},
  year={2025}, journal={arXiv preprint arXiv:2507.10524},
  doi={10.48550/arXiv.2507.10524}, url={https://arxiv.org/abs/2507.10524}
}

@inproceedings{chung2015gatedfeedback,
  title={Gated Feedback Recurrent Neural Networks},
  author={Chung, Junyoung and Gulcehre, Caglar and Cho, Kyunghyun and Bengio, Yoshua},
  year={2015}, booktitle={Proceedings of the 32nd International Conference on Machine Learning},
  volume={37}, pages={2067--2075},
  url={https://proceedings.mlr.press/v37/chung15.html}
}

@article{hwang2024fam,
  title={{TransformerFAM}: Feedback Attention is Working Memory},
  author={Hwang, Dongseong and Wang, Weiran and Huo, Zhuoyuan and Sim, Khe Chai and Moreno Mengibar, Pedro},
  year={2024}, journal={arXiv preprint arXiv:2404.09173},
  doi={10.48550/arXiv.2404.09173}, url={https://arxiv.org/abs/2404.09173}
}

@inproceedings{hoffmann2022training,
  title={Training Compute-Optimal Large Language Models},
  author={Hoffmann, Jordan and Borgeaud, Sebastian and Mensch, Arthur and Buchatskaya, Elena and Cai, Trevor and Rutherford, Eliza and Casas, Diego de Las and Hendricks, Lisa Anne and Welbl, Johannes and Clark, Aidan and others},
  booktitle={Advances in Neural Information Processing Systems},
  year={2022}, url={https://arxiv.org/abs/2203.15556}
}

@article{toshniwal2024openmathinstruct2,
  title={{OpenMathInstruct-2}: Accelerating {AI} for Math with Massive Open-Source Instruction Data},
  author={Toshniwal, Shubham and Du, Wei and Moshkov, Ivan and Kisacanin, Branislav and Ayrapetyan, Alexan and Gitman, Igor},
  year={2024}, journal={arXiv preprint arXiv:2410.01560},
  url={https://arxiv.org/abs/2410.01560}
}

@article{huang2026trading,
  title={Trading Depth for Time in Recurrent Transformers},
  author={Huang, Zeyi and He, Xuehai and Shen, Yelong and Lee, Yong Jae},
  journal={arXiv preprint arXiv:2609.21605},
  year={2026}
}

\clearpage
\appendix
\section{Full Numerical Results}
\label{sec:full-results}
Tables~\ref{tab:tf} and~\ref{tab:post} provide the full numerical results underlying
Figures~\ref{fig:pretraining} and~\ref{fig:posttraining},
including reference results and teacher scores. In Table~\ref{tab:tf}, the
TF-KD reference is a separate CE-only control trained from
scratch. This reference is not treated as a zero-sized teacher.

\paragraph{Computing BPB improvement.}
Bits per byte (BPB) expresses next-token negative log-likelihood in
bits, normalized by the number of bytes in evaluation text.
Let $b$ denote a student's validation BPB under TF-KD, and
$b^{\mathrm{ref}}$ the CE-only reference in the same row and training-condition
group of Table~\ref{tab:tf}. In Figure~\ref{fig:pretraining}, the unlabeled leftmost point
represents the TF-KD CE-only control.
The first panel shows absolute BPB; the second reports
relative reduction. Figure~\ref{fig:pretraining} additionally shows latency in its third
panel. Relative BPB improvement is computed as:
\begin{equation}
\operatorname{Improvement}\;(\%)
=100\times\frac{b^{\mathrm{ref}}-b}{b^{\mathrm{ref}}}.
\label{eq:bpb-improvement}
\end{equation}
Each architecture is normalized by its own condition-specific reference
BPB. TF-KD compares separate from-scratch runs.
A positive value indicates improvement, zero
indicates unchanged BPB, and a negative value indicates degradation; higher
is better. For example, the recurrent student with the 3.9B
teacher improves BPB from 0.877 to 0.851, a 2.96\% reduction.
Percentages are computed directly from the BPB
values reported in Table~\ref{tab:tf}, without intermediate rounding. This metric is
a relative BPB reduction, not a PPL reduction or an
absolute BPB difference. Each reference appears as the first horizontal
category, connected to the teacher-conditioned results, and has zero
relative improvement by definition. The absolute BPB panels share
one vertical scale; the improvement panels share another. Equally
spaced horizontal categories do not represent numerical teacher-size distances.
Displayed BPB values are rounded to three decimal places; plotted
improvements and differences computed from them inherit this precision limit.
\new{The evaluator follows this convention. The byte count includes
UTF-8 bytes of the target tokens; special tokens such as
the document-start token have zero bytes and are excluded from
both sums. Negative log-likelihood and bytes are summed over all
evaluated tokens before dividing (a micro-average, not a per-document mean).
Evaluation uses the first 10.5M tokens of the held-out shard,
in non-overlapping 2,048-token rows that each start at a document
boundary, so there is no sliding window or stride. Reported
values are from the final training step. Recurrent students
are evaluated token by token with sequential recurrent decoding.}

\begin{table*}[!ht]
\centering\small
\begin{tabular}{lrrrrrrr}
\toprule
Student & Reference & 0.8B & 1.3B & 2.0B
& 2.8B & 3.9B & Latency \\
\midrule
Teacher BPB & -- & 0.837 & 0.791 & 0.754
& 0.724 & 0.702 & -- \\
\midrule
\multicolumn{8}{l}{\emph{TF-KD}} \\
Vanilla 12 (0.43B) & 0.896 & 0.875 & 0.871 &
0.871 & 0.872 & 0.873 & x1.0 \\
Loop 2 & 0.885 & 0.864 & 0.861 & 0.861
& 0.861 & 0.862 & x1.9 \\
Loop 3 & 0.879 & 0.859 & 0.853 & 0.851
& 0.852 & 0.852 & x2.9 \\
Loop 4 & 0.873 & 0.856 & 0.850 & 0.848
& 0.849 & 0.849 & x3.9 \\
Recurrent & 0.877 & 0.860 & 0.853 & 0.852 &
0.851 & 0.851 & x1.1 \\
Recurrent + 1 thought & \textbf{0.865} & \textbf{0.846} & \textbf{0.840}
& \textbf{0.837} & \textbf{0.837} & \textbf{0.836} & x2.3 \\
\bottomrule
\end{tabular}
\caption{Pretraining validation BPB ($\downarrow$) under TF-KD.
Students use a 12-layer, 0.43B backbone.
Teacher sizes increase across columns.
Reference denotes the separately trained CE-only control. All TF-KD
runs, including the CE-only controls, train from scratch. Latency
is the decoding latency per generated token relative to the
12-layer vanilla student ($\times$1.0); it does not describe training
cost. Bold marks the lowest student
BPB in each column within each group; -- denotes unreported
values.}
\label{tab:tf}
\end{table*}

\begin{table*}[!ht]
\centering\small
\begin{tabular}{lrrrrrr}
\toprule
Student & No teacher & 0.8B & 1.3B & 2.0B
& 2.8B & 3.9B \\
\midrule
Teacher & -- & 40.1/25.0 & 43.6/31.8 & 48.5/31.8 &
50.2/34.2 & 51.8/34.8 \\
\midrule
\multicolumn{7}{l}{\emph{TF-KD}} \\
Vanilla & 31.8/23.6 & 34.4/26.2 & 34.8/26.2 & 34.0/27.6 &
34.4/27.8 & 34.4/26.4 \\
Loop 3 & 39.2/\textbf{26.8} & 39.6/28.0 & 41.8/29.2 & 41.2/29.0
& 42.4/29.2 & 41.8/29.4 \\
Recurrent & \textbf{40.5}/26.6 & 40.9/27.8 & 43.1/29.2 & 44.3/29.6 &
44.0/30.8 & 44.5/31.4 \\
\midrule
\multicolumn{7}{l}{\emph{OPD}} \\
Vanilla & 31.8/23.6 & 35.9/24.4 & 35.7/27.8 & 36.4/27.4 &
36.7/28.2 & 36.5/28.2 \\
Loop 3 & 39.2/\textbf{26.8} & 40.9/\textbf{28.2} & 42.8/\textbf{29.6} & 42.4/30.6
& 42.6/30.8 & 43.1/31.4 \\
Recurrent & \textbf{40.5}/26.6 & \textbf{44.0}/25.8 & \textbf{45.2}/29.4 & \textbf{46.2}/\textbf{32.2} &
\textbf{46.2}/\textbf{33.4} & \textbf{47.2}/\textbf{34.2} \\
\bottomrule
\end{tabular}
\caption{Post-training accuracy (\%, $\uparrow$), reported as GSM8K/MATH500, starting from the
project's own pretraining weights. Students retain the approximately 0.4B backbone
scale. The teacher row gives teacher accuracy; no-teacher entries are
the reported reference conditions. Bold marks the best student for
each metric in each column.}
\label{tab:post}
\end{table*}

\subsection{OPD from Pretrained Qwen2.5}
\label{sec:qwen-opd}
Table~\ref{tab:qwen-opd} reports results starting from pretrained
Qwen2.5-0.5B, rather than the project's own pretraining weights.
Both Loop 3 and Recurrent students are initialized from
pretrained Qwen2.5-0.5B weights and fine-tuned in their respective architectures.
\emph{No teacher} denotes fine-tuning without a teacher for all
three students. The teacher columns report OPD runs that
start from the same pretrained weights and use vanilla Qwen2.5
teachers.

\begin{table*}[!ht]
\centering\small
\begin{tabular}{lrrrr}
\toprule
Student & No teacher & Small teacher & Medium teacher
& Large teacher \\
& & (1.5B) & (3B) & (7B) \\
\midrule
Teacher & -- & 62.8/49.4 & 67.4/55.4 & 71.3/63.6 \\
\midrule
Vanilla & 49.3/36.2 & 52.5/36.8 & 52.2/38.6 & 53.7/37.6\\
Loop 3 & 51.0/36.2 & 53.5/37.2 & 54.0/39.2 & 55.2/39.4\\
Recurrent & 53.7/39.6 & 55.4/40.4 & 56.5/41.2 & 57.5/41.6\\
\bottomrule
\end{tabular}
\caption{Accuracy (\%, $\uparrow$), reported as GSM8K/MATH500.
In the \emph{No teacher} column, each student is pretrained
Qwen2.5-0.5B fine-tuned without a teacher in the Vanilla, Loop 3,
or Recurrent architecture. The remaining three columns report OPD results:
each student starts from pretrained Qwen2.5-0.5B, is fine-tuned on
the same data, and is distilled using a vanilla Qwen2.5
teacher of the indicated size (1.5B, 3B, or 7B).
The Teacher row reports teacher accuracy.}
\label{tab:qwen-opd}
\end{table*}

\section{Implementation Details of Student Computation}
\label{sec:impl}
This section gives the implementation details of the student computation
patterns summarized in Section~\ref{sec:method}.

\paragraph{Looped students.}
\new{The full 12-layer stack is applied $K$ times with
all weights shared. The residual stream is passed from one
loop to the next unchanged, with no normalization, projection, or
loop-index embedding in between. Every loop reuses the same rotary
positions, attention windows, per-layer residual scalars, and token-indexed value embeddings,
and re-adds the per-layer skip connection to the normalized token
embedding. The final RMSNorm and output head are applied once,
after the last loop. Each loop attends over its own
keys and values, so decoding keeps $12K$ KV-cache layer slots.}

\paragraph{Feedback injection.}
\new{The fed-back state $m_{t-1}$ is the output of
the final parameter-free RMSNorm, taken before the output head;
no further
normalization is applied when it re-enters. It is injected
at every layer $i$ in two ways. First,
$\gamma_i m_{t-1}$ is added to the residual stream $x$
before the block, where $\gamma_i$ is
a learnable scalar initialized to 0.1.
Second, per-layer bias-free linear maps project it into attention value
and key space, $v^{\mathrm{fb}}=W^v_i m_{t-1}$ and $k^{\mathrm{fb}}=W^k_i m_{t-1}$, which are
mixed with the local values and keys through per-head gates
$[g^{\mathrm{loc}},g^{\mathrm{fb}}]=2\sigma(G_i\,\mathrm{norm}(x))$:
\begin{align}
v&\leftarrow g^{\mathrm{loc}}v+g^{\mathrm{fb}}v^{\mathrm{fb}},\\
k&\leftarrow g^{\mathrm{loc}}k+g^{\mathrm{fb}}k^{\mathrm{fb}}.
\end{align}
$G_i$ is zero-initialized, so all gates start at one.
The state is zero at the first position of every
2,048-token training row and of every decoded sequence; it is
not reset at document boundaries inside a packed row.}

\paragraph{Recurrent training and decoding.}
\new{Training uses three parallel teacher-forced passes over
each sequence with shared weights. The first pass receives zero
feedback; each later pass receives the previous pass's final states
shifted right by one position. Gradients flow through all passes
(no detachment). Cross-entropy is applied at every position of every
pass and combined across passes (weights 0.1/0.3/0.6 in
the pretraining distillation runs for which we have records, and
equal weights for the post-training student's pretraining and fine-tuning),
whereas the KD term uses only the last pass. Sequential
decoding uses a single stream: one forward pass per token
with an ordinary KV cache and the previous token's
final state as feedback.
In post-training rollouts and evaluation, the prompt is also consumed
token by token in this recurrent mode, so there is
no separate parallel prefill. Pretraining BPB for recurrent students is
likewise evaluated token by token in this sequential recurrent mode,
not from the parallel multi-pass forward pass.}

\paragraph{Additional thinking step.}
\new{Each row interleaves 2,048 vocabulary tokens with 2,048 latent
positions (row length 4,096). A latent position has no token
embedding: its input is the fed-back final state of its
paired vocabulary token. It shares that token's rotary position and
value-embedding index, and attends in the same causal softmax as
vocabulary positions with shared attention weights but a separate MLP.
Feedback projections are shared across directions, while the gates and
residual scalars are separate for vocabulary-to-latent and latent-to-vocabulary feedback; latent
positions additionally receive the previous latent position's state as a
residual. Training runs one unsupervised vocabulary-only pass that seeds the
latent inputs, followed by three interleaved passes. The next-token cross-entropy
is read only at the latent position.}

\section{Experimental Details}
\label{sec:exp-details}
This section gives the experimental details summarized in the main
text.

\paragraph{Model family.}
\new{An $L$-layer model has hidden size $64L$ and $L/2$ attention
heads of dimension 128 (768 and 6 for the 12-layer
student), rotary positions, QK-normalized attention, parameter-free pre-RMSNorm, ReLU$^2$ MLPs,
untied input and output embeddings, a 2,048-token context, and a
32,768-token BPE vocabulary. Every layer also has a token-indexed value
embedding, which accounts for most parameters. Logged totals are 437.3M
for the 12-layer student (84.9M in Transformer blocks, 50.3M in
input and output embeddings, 302.0M in value embeddings) and 805.3M,
1.316B, 1.988B, 2.84B, and 3.892B for the 16-, 20-, 24-,
28-, and 32-layer teachers. Looped students add no parameters. The
recurrent student has 451.6M: 14.2M feedback projections (3.2\% overhead)
plus gates. The 14-layer vanilla reference has hidden size 896,
7 heads, and 604.6M parameters.}

\paragraph{Pretraining data and teacher budgets.}
\new{We use the shuffled 100B-token FineWeb-Edu release, split into 1,823
shards; the last shard is held out. Text is tokenized
with a 32,768-token byte-level BPE tokenizer trained on the same
corpus, and documents are packed into 2,048-token rows that each
start at a document boundary. Teachers are trained on 10
tokens per scaling parameter (Transformer-block and output-embedding parameters), which gives
2.35B, 4.35B, 7.30B, 11.38B, and 16.78B training tokens (4,480, 8,300,
13,920, 21,700, and 32,000 steps) for the 16-, 20-, 24-,
28-, and 32-layer teachers.}

Teachers are vanilla autoregressive models pretrained from scratch on
FineWeb-Edu \citep{penedo2024fineweb} using next-token CE. We use a fixed
data-to-parameter ratio in the spirit of compute-optimal scaling
\citep{hoffmann2022training}, rather than replicating the size-by-token-budget sweep of
\citet{lu2026strong}. Teacher size and pretraining token budget vary jointly,
so this sweep does not isolate teacher parameter count.
Teacher and student runs sample the same data pool
with different seeds, without repeating data within a run.

\paragraph{Teacher-forced KD protocol.}
Students train from scratch for 2,520 steps, processing 1.32B
tokens, with frozen teachers and the mixed objective in
Equation~\ref{eq:tf}. The student token budget follows the same
compute-optimal scaling motivation described above \citep{hoffmann2022training} and is
held fixed across student architectures and their CE-only controls.
CE and KD have fixed weights of 0.5/0.5,
rather than being swept as in \citet{lu2026strong}. KD is
token-averaged forward KL (teacher to student) over the full
vocabulary at temperature $\tau=1$; both losses use real-text histories.
The 14-layer vanilla reference uses the same training budget
and objective as the 12-layer students, but lies outside
the fixed-backbone comparison. Fixed backbone parameters do not imply
equal total parameters when feedback projections are included.

\paragraph{Pretraining optimization.}
\new{Runs use 524,288 tokens (256 sequences) per step in bf16
on up to 8 H100 GPUs. Transformer matrices are trained
with Muon (learning rate 0.02; momentum warmed from 0.85 to
0.95 over 300 steps; weight decay $0.2\,(12/L)^2$, decayed linearly
to zero),
and embeddings, the output head, and scalars with AdamW ($(\beta_1,\beta_2)=(0.8,0.95)$;
learning rates 0.3, 0.004, and 0.5, scaled by $(64L/768)^{-1/2}$). There
is no warm-up and no gradient clipping. The learning rate
is constant for the first half of training and decays
linearly to zero over the second half.}
For TF-KD students, decay therefore starts at step 1,260.

\paragraph{Differences from the reference framework.}
We follow the pretraining distillation framework of \citet{lu2026strong},
but do not inherit their optimizer and context settings.
They use an 8,192-token context, a global batch of
256 sequences, AdamW with $(\beta_1,\beta_2)=(0.9,0.95)$, weight decay 0.1,
gradient clipping 1.0, and bf16. Their peak learning rate
is $3\times10^{-4}$, with 5\% linear warm-up and cosine decay
to 10\% of peak. Our settings are given above.


\paragraph{Post-training starting checkpoints.}
\new{The pretraining experiments (Table~\ref{tab:tf}) train the 12-layer students and all
teachers with 10 training tokens per scaling parameter, which is
2,520 steps for the students. For post-training we do not
reuse these weights. To give post-training a stronger starting point,
every student and teacher is pretrained again from scratch with
20 tokens per scaling parameter, using cross-entropy only: 5,040 steps
(2.64B tokens) for the students, and twice the token budgets
listed above for the teachers. Each teacher is then fine-tuned
on the post-training data without a teacher. Student fine-tuning, TF-KD,
and OPD all start directly from the student's pretraining checkpoint:
there is no separate fine-tuning stage before distillation, and in
OPD the teacher-forced and on-policy steps alternate within a single
run.}

\paragraph{Post-training data.}
\new{OpenMathInstruct-2 problems and solutions are synthesized from the GSM8K and
MATH training sets. After shuffling the \texttt{train\_5M} split (seed 42),
we hold out 4,096 examples and train on the remaining
4,995,904 problem--solution pairs. We ran no additional decontamination check
against the evaluation sets. The post-training teachers are pretrained
on 4.70B, 8.70B,
14.60B, 22.76B, and 33.56B tokens (16- to 32-layer) before fine-tuning.}

\paragraph{Post-training optimization and objectives.}
\new{Each run starts a fresh optimizer of the pretraining type,
with learning rates set to 0.2 times the pretraining values,
3\% linear warm-up, a constant phase, and linear decay to
zero over the last half; weight decay is zero. Conversations
are rendered in a chat template and packed into 2,048-token
rows, and all losses are restricted to assistant tokens. In
TF-KD, the forward KL is computed over the full vocabulary
at temperature 1 and averaged over assistant tokens, with the
fine-tuned teacher run on the same batch. OPD alternates
two kinds of steps in equal numbers. A teacher-forced step
uses the TF-KD loss $0.5\,\mathcal{L}_{\mathrm{CE}}+0.5\,\mathrm{KL}(p_T\|p_S)$ on the reference solutions.
An on-policy step uses $0.5\,\mathcal{L}_{\mathrm{CE}}+0.5\,\mathrm{KL}(p_S\|p_T)$, with the cross-entropy on
the reference solutions and the reverse KL on student rollouts;
it has no forward-KL term.
For the on-policy term, 64
training questions per GPU (at most 1,024 prompt tokens) are
sampled on every second step, about 1.2M rollouts per run;
the student generates an answer of up to 512 tokens
at temperature 1, stopping at the end-of-answer token, and the
reverse KL is averaged over the sampled answer tokens. For
the recurrent student, rollouts use recurrent decoding, and the distillation
term of either step is applied to the last pass
of the parallel forward pass.}

\paragraph{Evaluation protocol.}
\new{Each problem is presented as a single user turn
in the chat template. The model generates a response
greedily without a calculator tool until the end-of-answer token.
GSM8K allows up to 192 new tokens and uses
numeric exact match on the first ``\#\#\#\# $N$'' or
``answer is
$N$'' pattern, falling back to the last number. For MATH500,
``Put your final answer in \textbackslash boxed\{\}.'' is appended
to the
problem and up to 512 new tokens are generated. Outputs
that reach the token limit are graded as they are
(about 35--40\% of outputs).}

\paragraph{Latency measurement.}
\new{Decoding latency in Table~\ref{tab:tf} and
Figure~\ref{fig:pretraining} is the wall-clock time per generated
token with KV caching, measured on a single NVIDIA H100
(80\,GB) at batch size 1 with a 1,024-token prompt
and 128 greedily
decoded tokens. Prompt prefill is timed separately and excluded.
Models run in PyTorch eager mode (no compilation) with bfloat16
autocast, and the GPU is synchronized before each timestamp. Latency
does not depend on weight values, so models are
randomly initialized
from the training configurations. All students are loaded in one
process on the same GPU and timed in rotating order;
after one warm-up round we take the median over six
rounds and
report it relative to the 12-layer vanilla student.}

\paragraph{Qwen2.5 fine-tuning and OPD.}
\new{Problems are rendered in the ChatML template and the
loss is computed on assistant tokens only, with a 2,048-token
context, 524,288 tokens per step, and 4,096 held-out problems for
validation. All runs use AdamW ($\beta=(0.9,0.95)$, gradient clipping at
1.0, no weight decay) with a 3\% linear warm-up and
a linear decay to zero over the last half of
training. The learning rate is $2\times10^{-5}$ for pretrained weights, $10^{-4}$
for the new feedback projections, and $2\times10^{-5}$ for the residual
feedback scales. For the recurrent student, the fed-back state is
normalized to unit RMS before injection, cross-entropy is averaged over
the three passes with equal weights, the distillation term uses
the last pass, and rollouts use recurrent decoding. OPD alternates
two kinds of steps in equal numbers: a teacher-forced step
with $0.5\,\mathcal{L}_{\mathrm{CE}}+0.5\,\mathrm{KL}(p_T\|p_S)$ on reference solutions, and an on-policy step
with $0.5\,\mathcal{L}_{\mathrm{CE}}+0.5\,\mathrm{KL}(p_S\|p_T)$, where the reverse KL is on
student rollouts; rollouts are
sampled at temperature 1 with at most 512 tokens. The
1.5B, 3B, and 7B teachers are vanilla Qwen2.5 models fully
fine-tuned on the same data for one epoch with a
learning rate $10^{-5}$. Evaluation follows the protocol above.}

\paragraph{Converting pretrained Qwen2.5.}
\new{\emph{Recurrent.} Let $m_{t-1}$ be the final-layer hidden state of
the previous token after the final normalization, rescaled to unit
RMS, with $m_{0}=0$. The pretrained weights are kept and
every layer $i$ receives $m_{t-1}$ through four added terms.
Before the block, the residual stream becomes
$x \leftarrow x+\lambda_i m_{t-1}+s_i R_i m_{t-1}$,
where $\lambda_i$ is a learned scalar,
$R_i$ a learned $d\times d$ matrix, and $s_i$ a fixed
scale equal to the typical magnitude of the pretrained residual
stream entering layer $i$. Inside attention, linear maps of $m_{t-1}$
are added to the query, key, and value projections before
the rotary embedding:
$q\leftarrow q+Q_i m_{t-1}$, $k\leftarrow k+K_i m_{t-1}$, $v\leftarrow v+V_i m_{t-1}$.
Training uses three parallel passes as in Section~\ref{sec:impl}: each pass
reads the previous pass's states shifted right by one position,
without stopping gradients. All new feedback parameters
($\lambda_i$, $R_i$, $Q_i$, $K_i$,
$V_i$) are zero-initialized, so the
converted model reproduces pretrained Qwen2.5-0.5B exactly before training (validation BPB
0.2493 for both). The final hidden state of Qwen2.5-0.5B has
an RMS of about 8, against 0.015 for the
input embeddings. Without normalizing the fed-back state, and with a
learning rate of $10^{-3}$ for the new parameters, the loss
of the second and third passes diverged within two steps;
we therefore normalize the state to unit RMS and use
$10^{-4}$. The first training pass receives zero feedback and is
thus the original computation.
\emph{Loop~3.} The output of the 24-layer stack is fed back
as its input twice; rotary positions are shared across loops,
and the final normalization and the loss are applied once,
after the last loop. This model is not equal to
the pretrained model at initialization (validation BPB 3.00 at step
0) and falls below the pretrained model's BPB within the
first 100 steps.
\emph{Cost.} Without a teacher, on 8 H100 GPUs, one training
step takes about 1.65\,s for Vanilla, 2.8\,s for Loop~3, and
3.85\,s for Recurrent, so one epoch takes about 2.0, 3.3,
and 4.7 hours. Recurrent decoding uses one forward pass per
token and Loop~3 uses three.}

\subsection{Evaluation and Reporting Conventions}
\label{sec:reporting-conventions}
GSM8K evaluation uses the 1,319 test questions
\citep{cobbe2021gsm8k}. MATH500 evaluation uses the 500 problems
in the public \texttt{HuggingFaceH4/MATH-500} test file
\citep{hendrycks2021math,lightman2023verify}, with Math-Verify equivalence grading.
Both benchmarks are used only for evaluation in the
mathematical post-training experiments.

Reported accuracy and BPB values are from single runs,
with one seed per configuration, rather than averages or
maxima across seeds. We report no confidence intervals or
across-seed variability. Latency instead reports the median of six
timing rounds after one warm-up round, as described above.

The mathematical post-training experiments use the final checkpoint,
not the checkpoint with the highest test accuracy.
The teacher-forced pretraining KD objective fixes CE and KD
weights at 0.5/0.5 rather than sweeping them. The
feedback-normalization and learning-rate adjustment for Qwen2.5 is described
above; these settings should not be interpreted as an
exhaustive hyperparameter search.

\paragraph{Scope of reported computational costs.}
The reported Qwen2.5 no-teacher epoch times on eight H100
GPUs correspond to approximately 16, 26.4, and 37.6
GPU-hours for Vanilla, Loop~3, and Recurrent, respectively.
These are per-run estimates, not an aggregate project budget.
They exclude teacher training, OPD teacher computation, other training
runs, and exploratory or unsuccessful runs. The per-run timings
and token budgets do not establish matched training compute.

\subsection{Artifact Sources, Licenses, and Data Scope}
\label{sec:artifact-documentation}
\paragraph{Data sources and coverage.}
Pretraining uses the 100B-token sample of FineWeb-Edu
\citep{penedo2024fineweb}, an English educational web-text corpus. Its
official dataset card documents its source and educational filtering.
\footnote{\url{https://huggingface.co/datasets/HuggingFaceFW/fineweb-edu}}
Post-training uses the \texttt{train\_5M} split of OpenMathInstruct-2
\citep{toshniwal2024openmathinstruct2}, an English mathematical instruction dataset
with synthetic problems and solutions derived from GSM8K and
MATH training problems.
\footnote{\url{https://huggingface.co/datasets/nvidia/OpenMathInstruct-2}}
GSM8K covers grade-school mathematical word problems; MATH500 covers
competition-style mathematics. Both evaluation datasets are in English.
We use the public MATH500 test subset, not the
PRM800K process-supervision annotations.
\footnote{\url{https://huggingface.co/datasets/HuggingFaceH4/MATH-500}}
These datasets do not establish demographic representativeness, and our
evaluations do not assess performance across demographic groups or
languages beyond English.

\paragraph{Dataset licenses and terms.}
FineWeb-Edu is released under Open Data Commons Attribution
(ODC-By) v1.0 and is also subject to Common Crawl's
Terms of Use, as specified in its dataset card.
OpenMathInstruct-2 is released under Creative Commons Attribution 4.0
(CC-BY-4.0). The official GSM8K repository carries the MIT license.
\footnote{\url{https://github.com/openai/grade-school-math}}
The MATH and PRM800K source repositories carry MIT licenses.
\footnote{\url{https://github.com/hendrycks/math};
\url{https://github.com/openai/prm800k}}
However, the MATH500 dataset card does not specify a
separate dataset license. Repository licenses should not be interpreted
as establishing unrestricted rights to underlying third-party problem text.

\paragraph{Pretrained models and evaluation software.}
The adaptation experiments use base Qwen2.5 models
\citep{qwen2024report}: a 0.5B student and 1.5B, 3B, and
7B teachers, each fine-tuned as described above. The official
Qwen2.5-0.5B, Qwen2.5-1.5B, and Qwen2.5-7B model cards specify Apache-2.0.
\footnote{\url{https://huggingface.co/Qwen/Qwen2.5-0.5B};
\url{https://huggingface.co/Qwen/Qwen2.5-1.5B};
\url{https://huggingface.co/Qwen/Qwen2.5-7B}}
Qwen2.5-3B instead uses the Qwen Research License Agreement,
which permits non-commercial research and evaluation and requires a
separate license for commercial use. Its attribution and redistribution
conditions also apply to relevant derivatives and distributed models
improved using its outputs or results.
\footnote{\url{https://huggingface.co/Qwen/Qwen2.5-3B/blob/main/LICENSE}}
These experiments study training and evaluation, not production deployment.
The official base-model cards recommend post-training rather than using
base models directly for conversations. Math-Verify, used for MATH500
equivalence grading, is released by Hugging Face under Apache-2.0.
\footnote{\url{https://github.com/huggingface/Math-Verify}}

\paragraph{Privacy and offensive-content limitations.}
FineWeb-Edu's dataset card reports upstream URL-level filtering to
reduce toxic and offensive content, but explicitly warns that
harmful content and web-derived biases may remain. Educational filtering
is not a guarantee of privacy protection or harmlessness.
The use of public datasets does not establish that
they contain no personally identifying information. Our reported language-modeling
and mathematical evaluations do not constitute a privacy or
offensive-content audit, and we make no claim of complete
anonymization or removal of harmful content.

\section{AI Assistance}
\label{sec:ai-assistance}
We used GPT-6.1 Sol and Opus 5.5 to assist
with experimental coding and manuscript writing.

\end{document}